\documentclass[letterpaper, 10 pt, conference]{ieeeconf}  
\IEEEoverridecommandlockouts                           
\usepackage{graphics} 
\usepackage{epsfig} 
\usepackage{mathptmx} 
\usepackage{times} 
\usepackage{amsmath} 
\usepackage{amssymb}  
\usepackage{xcolor}
\usepackage{array, multirow,graphicx}
\usepackage{float}
\usepackage{graphicx}
\usepackage{subfigure}
\let\labelindent\relax
\usepackage{enumitem}
\usepackage{lipsum}
\usepackage{url}
\usepackage{hyperref}
\hypersetup{
    colorlinks=true,
    linkcolor=black,
    citecolor=black,
    filecolor=black,
    urlcolor=gray,
    breaklinks=true,
}

\makeatletter
\renewcommand{\@makefntext}[1]{%
  \parindent 0pt%
  \noindent\hbox{\@makefnmark}~#1%
}
\makeatother

\title{\LARGE \bf
Toward Human-Robot Teaming: Learning Handover Behaviors from 3D Scenes
}

\author{Yuekun Wu, Yik Lung Pang, Andrea Cavallaro, Changjae Oh%
\thanks{This work was supported in part by the CHIST-ERA program through the project CORSMAL, under the UK EPSRC Grant EP/S031715/1 and in part by the Royal Society Research Grant RGS\textbackslash R2\textbackslash242051. 
This research utilised the Sulis Tier 2 HPC platform under the UK EPSRC Grant EP/T022108/1 and the HPC Midlands+ consortium.
}
\thanks{Yuekun Wu, Yik Lung Pang, and Changjae Oh are with Centre for Intelligent Sensing, Queen Mary University of London, UK.
{\tt\small \{yuekun.wu,y.l.pang,c.oh\}@qmul.ac.uk}. Andrea Cavallaro is with Idiap Research Institute and École Polytechnique Fédérale de Lausanne, Switzerland.
{\tt\small andrea.cavallaro@epfl.ch}.}%
}
\begin{document}
\maketitle
\thispagestyle{empty}
\pagestyle{empty}
\begin{abstract}
Human-robot teaming (HRT) systems often rely on large-scale datasets of human and robot interactions, especially for close-proximity collaboration tasks such as human-robot handovers. 
Learning robot manipulation policies from raw, real-world image data requires a large number of robot-action trials in the physical environment. 
Although simulation training offers a cost-effective alternative, the visual domain gap between simulation and robot workspace remains a major limitation. 
We introduce a method for training HRT policies, focusing on human-to-robot handovers, solely from RGB images without the need for real-robot training or real-robot data collection.
The goal is to enable the robot to reliably receive objects from a human with stable grasping while avoiding collisions with the human hand.
The proposed policy learner leverages sparse-view Gaussian Splatting reconstruction of human-to-robot handover scenes to generate robot demonstrations containing image-action pairs captured with a camera mounted on the robot gripper. As a result, the simulated camera pose changes in the reconstructed scene can be directly translated into gripper pose changes. Experiments in both Gaussian Splatting reconstructed scene and real-world human-to-robot handover experiments demonstrate that our method serves as a new and effective representation for the human-to-robot handover task, contributing to more seamless and robust HRT. 
\end{abstract}

\section{Introduction}
Designing Human-Robot Teaming (HRT) systems requires enabling robotic systems that can interpret and respond to human actions in shared workspaces~\cite{simoes2022designing}. 
For tasks such as human-to-robot object handovers, training robot manipulation policies involves collecting large volumes of demonstration data involving both human and robot motion, which raises practical challenges in terms of cost, time, and safety~\cite{ohpl,wyk}.
To address these limitations, we propose a system for training robot handover policies without any real-robot movement or sim-to-real adaptation~\cite{sim2real}. Our method reconstructs 3D scenes of handover interactions from sparse RGB-D viewpoints using Gaussian Splatting (GS)~\cite{kerbl20233d}. We then simulate robot viewpoints by changing the camera pose in the reconstructed 3D scene, creating a sequence of synthetic hand-eye observations aligned with corresponding robot motion commands. These serve as demonstrations for supervised policy learning for reaching and grasping tasks entirely in simulation.

\begin{figure}[t]
  \vspace{1 mm}
  \centering
  \subfigure[3D reconstruction]{
    \includegraphics[width=0.45\linewidth]{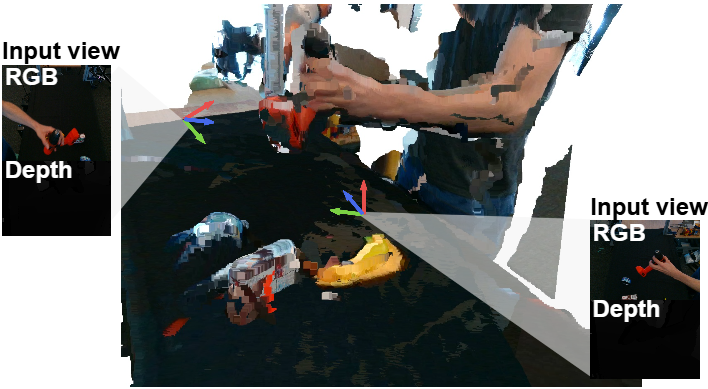}
  }
  \hspace{0.001\linewidth}
  \subfigure[Generating demonstrations]{
    \includegraphics[width=0.45\linewidth]{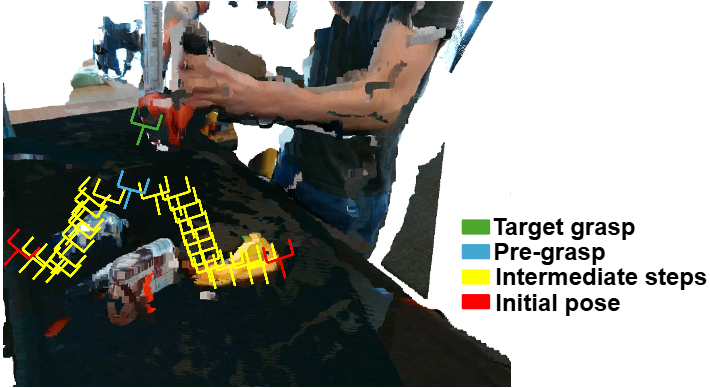}
  }\\
  \hspace{0.01\linewidth}
  \subfigure[Handover demonstrations]{
    \includegraphics[height=0.29\linewidth]{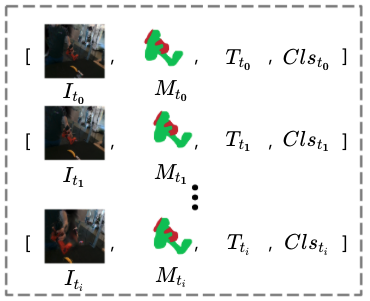}
    }
  \hspace{0.001\linewidth}
  \subfigure[Handover policy training]{
    \includegraphics[height=0.29\linewidth]{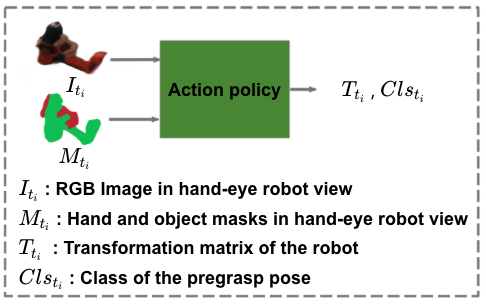}
    }
  \caption{Overview of our method. (a) Given sparse-view RGB-D handover images, we reconstruct a 3D scene using Gaussian Splatting (GS) and then estimate grasp poses from the object and hand point clouds extracted from the GS scene. (b) We then use the GS scene and grasp pose to generate the gripper's trajectory toward the pre-grasp pose and the hand-eye image at each sampled pose. (c) Each trajectory becomes a handover demonstration dataset that includes hand-eye images, object and hand masks, transformations of the gripper pose, and pre-grasp pose labels. (d) The dataset is used to train a handover policy. For inference, only the hand-eye RGB image and masks are required.}
    \label{fig1}
\end{figure}
\section{Method}
\subsection{Grasp pose estimation} 
Fig.~\ref{fig1} shows the main steps of the proposed method. We leverage 3D Gaussian Splatting (GS) to reconstruct human-object interaction scenes from sparse-view RGB-D inputs~\cite{zhu2024fsgs}, providing realistic, metric-scale environments for sim-to-real robot training. To consider a human hand when grasping hand-held objects, we extract object and hand point clouds and estimate diverse grasp candidates using GraspNet~\cite{mousavian20196}. Unsafe grasps, those colliding with the human hand, are then filtered by proximity checks in the grasp frame.
For effective transfer to the robot, grasp poses are aligned with the reconstructed scene by adjusting coordinate offsets, which facilitates direct mapping from camera to gripper motion. This ensures consistency between perception and control, critical for collaborative physical interaction.

\subsection{Handover demonstrations dataset} 
To support flexible handover strategies under human guidance, we simulate multiple handover demonstrations per grasp. Initial robot poses are sampled spherically around the grasp, with filtering to ensure realistic robot-reaching trajectories that consider human-object geometry and safety zones.

The robot reaching task towards the object is performed in three phases. In phase 1, the robot arm is moved to align the gripper's reaching direction with the object by locating it at the center of the image. In phase 2, the robot adjusts only its position via linear interpolation until it is within a pre-defined distance to the pre-grasp pose. In phase 3, both position and orientation are refined simultaneously, using linear interpolation for position and spherical linear interpolation for rotation, to reach the final pre-grasp pose.

We then project the hand and object into the camera view using the 6D hand-eye pose, and render RGB and mask images for each timestep. The constructed dataset is then used for visuomotor policy learning.

\subsection{Policy learning for human-to-robot handover}
Our policy network learns to control the robot’s 6-DoF motion and execute grasping from simulated hand-eye images and binary masks. At each step, the policy predicts a pose update and a grasp probability, enabling adaptive handover timing.
The policy is trained with a multi-objective loss combining translation, rotation, and grasp classification, allowing the system to generalize across object types and hand poses. 

\section{Experiment}
To evaluate the effectiveness of our method in human-robot teaming scenarios under human guidance, we perform experiments in both simulated 3D scenes and real-world robot handover tasks. In simulation, we assess how well a policy learned from RGB images and hand-object masks guides the robot to a safe, effective pre-grasp pose. On real hardware, we validate transferability and execution efficiency during human-to-robot handovers involving diverse objects and participants.
\subsection{Simulated handover in 3D scenes}

\begin{figure}[t]
\centering
\includegraphics[width=0.98\linewidth]{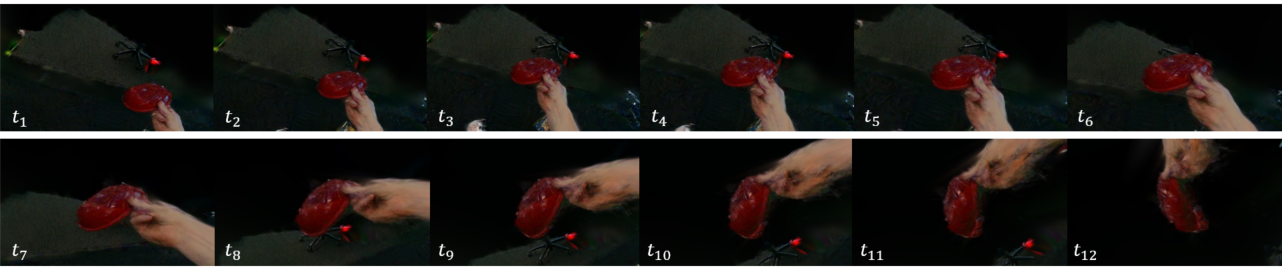}\\
\footnotesize (a) \\
\includegraphics[width=0.98\linewidth]{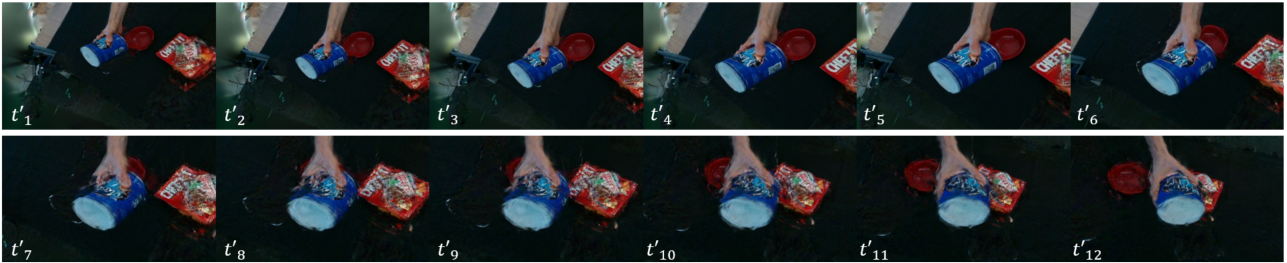}
\footnotesize (b) \\
\vspace{-5pt}
\caption{3D simulation trajectories for (a) seen and (b) unseen objects.}
\label{render-vis}
\end{figure}

In simulation, the robot predicts actions based on RGB input until reaching a pre-grasp state or timeout.
Fig.~\ref{render-vis} illustrates simulated trajectories toward the pre-grasp pose. Across both seen and unseen objects, the robot consistently orients, approaches, and adjusts its pose to enable safe and effective handover. This behavior reflects the model’s capacity to autonomously execute human-aware reaching strategies aligned with intended trajectory design—crucial for seamless human-robot collaboration.
The robot consistently maintains appropriate distances, aligns with object centers, and avoids the human hand, demonstrating its capability to plan goal-directed actions under autonomous control. 

\subsection{Real-World handover under human guidance}
\begin{figure}[t]
  \centering
  \includegraphics[width=0.98\linewidth]{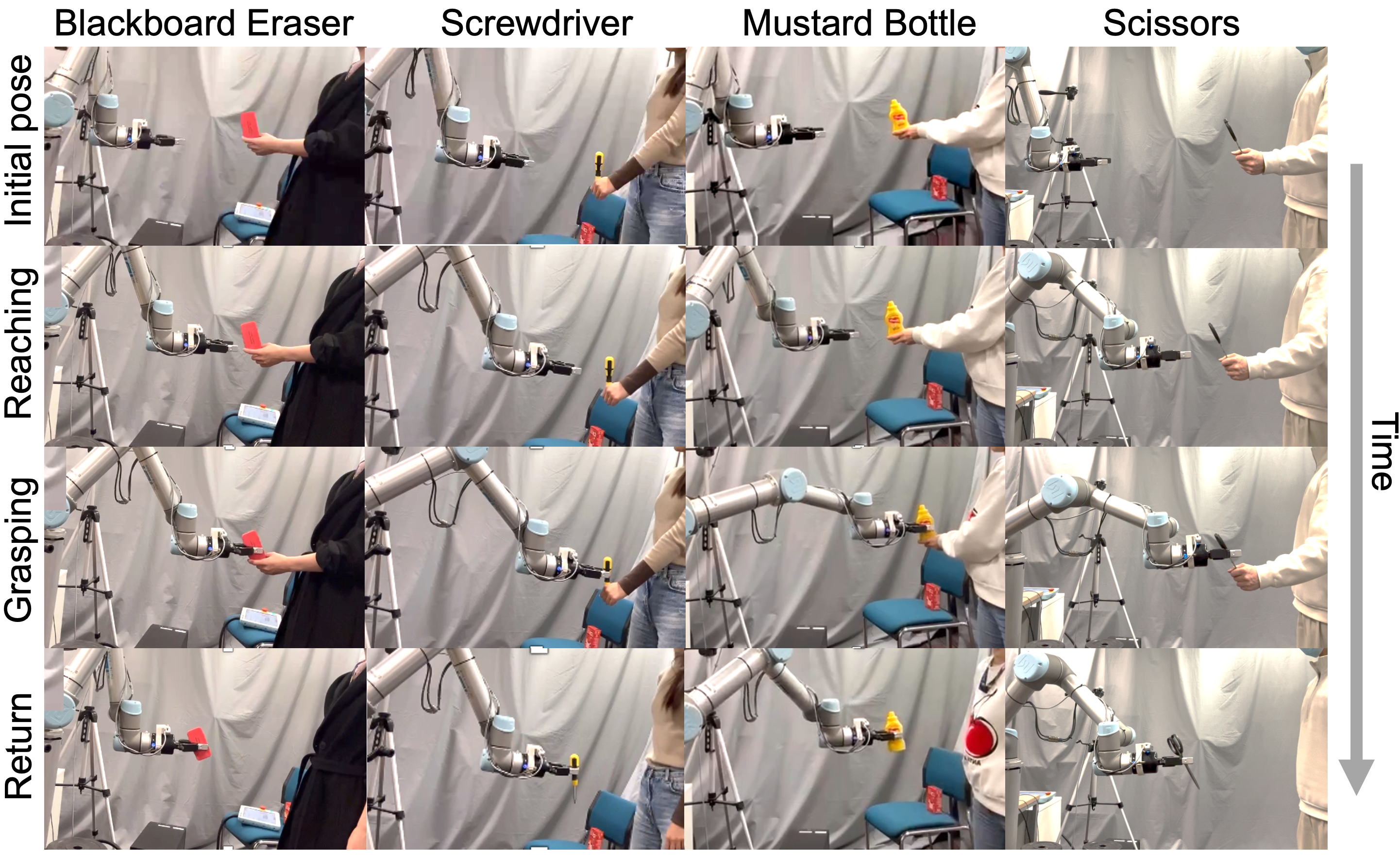}
  \caption{Examples of successful human-to-robot handovers of household objects using the proposed method with RGB image and hand-object mask.}
  \label{fig5}
  \vspace{-9pt}
  \vspace{-9pt}
\end{figure}

We conducted handovers involving 6 objects and 4 participants. In Fig.~\ref{fig5}, we showed that our method can perform human-to-robot handovers on general household objects. As observed from the results, our model exhibits robust and effective handover performance for both seen and unseen objects.
Our system represents a shift toward semi-autonomous robot teammates: it interprets high-level visual cues, executes safe grasp plans, and remains adaptable to new objects—all without requiring hand-engineered robot control. This ability to fluidly switch control modes and collaborate naturally supports the broader goal of trustworthy human-robot teaming under human guidance.

\section{Conclusion}
We introduced a novel framework for training human-to-robot handover policies without real-robot movement. Our system enables efficient, safe learning for HRT scenarios by simulating visual experiences through 3D Gaussian Splatting. This eliminates the need for costly motion data collection and supports scalable robot learning. In future work, we aim to integrate richer human intention understanding and use foundation models to expand interaction generalization.


{
\bibliographystyle{IEEEtran} 
\bibliography{main}
}

\end{document}